\documentclass[conference]{IEEEtran}
\IEEEoverridecommandlockouts
\usepackage{cite}

\usepackage{float}
\usepackage{url}

\usepackage{subcaption}
\usepackage{amsmath,amssymb,amsfonts}
\usepackage{algorithmic}
\usepackage{graphicx}
\usepackage{textcomp}
\usepackage{xcolor}
\usepackage{soul}
\usepackage{tcolorbox}
\sethlcolor{cyan}
\def\BibTeX{{\rm B\kern-.05em{\sc i\kern-.025em b}\kern-.08em
    T\kern-.1667em\lower.7ex\hbox{E}\kern-.125emX}}
\begin{document}

\title{A Decade of Wheat Mapping for Lebanon\\
\thanks{*Corresponding Author: aghandour@cnrs.edu.lb.}
}

\author{\IEEEauthorblockN{Hasan Wehbi†}
\IEEEauthorblockA{\textit{Earth Observation Department} \\
\textit{RASID SARL}\\
Beirut, Lebanon}
\and
\IEEEauthorblockN{Hasan Nasrallah†}
\IEEEauthorblockA{\textit{Earth Observation Department} \\
\textit{RASID SARL}\\
Beirut, Lebanon}
\and
\IEEEauthorblockN{Mohamad Hasan Zahweh}
\IEEEauthorblockA{\textit{Faculty of Engineering} \\
\textit{Lebanese University}\\
Beirut, Lebanon}
\and
\IEEEauthorblockN{Zeinab Takach}
\IEEEauthorblockA{\textit{Electrical and Computer Engineering Department} \\
\textit{American University of Beirut}\\
Beirut, Lebanon}
\and
\IEEEauthorblockN{Veera Ganesh Yalla}
\IEEEauthorblockA{\textit{IHub Data} \\
\textit{IIIT Hyderabad }\\
Hyderabad, India}
\and
\IEEEauthorblockN{Ali J. Ghandour*}
\IEEEauthorblockA{\textit{National Center for Remote Sensing} \\
\textit{National Council for Scientific Research}\\
Beirut, Lebanon}
}

\maketitle
\def\thefootnote{†}\footnotetext{These authors contributed equally to this work}\def\thefootnote{\arabic{footnote}}

\begin{abstract}
Wheat accounts for approximately 20\% of the world's caloric intake, making it a vital component of global food security. Given this importance, mapping wheat fields plays a crucial role in enabling various stakeholders, including policy makers, researchers, and agricultural organizations, to make informed decisions regarding food security, supply chain management, and resource allocation. 
In this paper, we tackle the problem of accurately mapping wheat fields out of satellite images by introducing an improved pipeline for winter wheat segmentation, as well as presenting a case study on a decade-long analysis of wheat mapping in Lebanon. We integrate a Temporal Spatial Vision Transformer (TSViT) with Parameter-Efficient Fine Tuning (PEFT) and a novel post-processing pipeline based on the Fields of The World (FTW) framework. Our proposed pipeline addresses key challenges encountered in existing approaches, such as the clustering of small agricultural parcels in a single large field. By merging wheat segmentation with precise field boundary extraction, our method produces geometrically coherent and semantically rich maps that enable us to perform in-depth analysis such as tracking crop rotation pattern over years. Extensive evaluations demonstrate improved boundary delineation and field-level precision, establishing the potential of the proposed framework in operational agricultural monitoring and historical trend analysis. By allowing for accurate mapping of wheat fields, this work lays the foundation for a range of critical studies and future advances, including crop monitoring and yield estimation.
\end{abstract}


\section{Introduction}
Accurate and long-term crop mapping is critical for agricultural monitoring, food security assessments, and policy formulation. In regions where agricultural parcels vary widely in size, traditional pixel-based segmentation methods often lack the precise field boundaries required for operational applications. Our previous work \cite{b1} employed a Temporal Spatial Vision Transformer (TSViT) combined with Parameter-Efficient Fine Tuning (PEFT) for winter wheat segmentation, demonstrating promising results using weakly supervised learning. However, medium spatial resolution and sparse label availability led to the following limitations: 
\begin{itemize}
\item \textbf{Inaccurate Field Boundaries and Out-of-Bounds Predictions:} The model struggled to delineate true wheat field boundaries, often resulting in under-segmentation (missing parts of fields) or over-segmentation (including non-wheat areas). This inaccuracy was particularly pronounced in areas with heterogeneous land cover or where wheat fields were adjacent to other types of vegetation.
\item \textbf{Merging of Adjacent Wheat Fields:} The model frequently failed to distinguish between closely spaced wheat fields, merging them into a single larger field. This issue arose due to the limited spatial resolution.
\item \textbf{High Noise in Segmentation Results:} The segmentation results exhibited significant noise, with scattered pixels or small regions incorrectly classified as wheat.
\end{itemize}
\begin{figure}[h]
    \centering
    \includegraphics[width=0.9\linewidth]{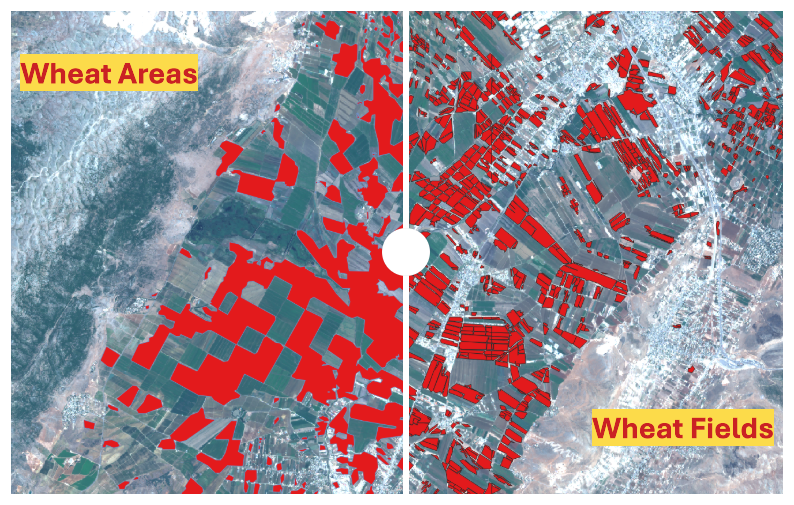}
    \caption{Slider image showing wheat areas: (left) detected using our conventional model and (right) accurately delineated by our newly proposed model.}
    \label{fig:compare}
\end{figure}

These limitations would reduce the accuracy of any crop field-related statistics. Hence, this work presents an improved processing pipeline that incorporates a dedicated field delineation model based on the open-source \texttt{ftw-baseline} \cite{b2}. This work not only builds on our earlier methodology but also provides a refined approach to reconcile pixel-level predictions with the geometric realities of agricultural fields. Our contribution in this paper is three-fold:
\begin{itemize}
    \item \textbf{Accurate Parcel-level Wheat Mapping:} Use precise field delineation to improve wheat mapping results.
    \item \textbf{Refined Post-Processing:} A set of post-processing steps to improve field delineation results.
    \item \textbf{A Decade-Long Analysis of Wheat Mapping in Lebanon:} A comprehensive case study evaluating wheat field trends over a ten-year period, with actionable statistics for policy makers and stakeholders to act on.
    \end{itemize}

Figure \ref{fig:compare} shows that the proposed pipeline overcomes the stated challenges, allowing more reliable agricultural monitoring. An interactive version of this Figure can be accessed here: https://shorturl.at/PONJH
    
\section{Previous Work}
Several studies have explored crop segmentation using remote sensing data. For example, \cite{b9} experimented with various approaches in Africa, including 3D U-Net and ConvLSTM architectures, while handling different input types such as Sentinel-1, Sentinel-2, and Planet imagery. However, this approach was outperformed on the PASTIS dataset \cite{b6}, which used only Sentinel-2 data, but incorporated more advanced models like CNNs with temporal attention \cite{b9} or temporal-spatial transformers \cite{b7}. These methods typically rely on around 70 images per year, resulting in high memory requirements. In contrast, \cite{b12} demonstrated that competitive results could be achieved using only 12 images per year, through a streamlined and efficient pipeline.

In the field delineation domain, earlier work using Sentinel-2 data was relatively limited to \cite{b2}. The manuscripts \cite{b13} and \cite{b14} used the conventional UNet approach. The authors in \cite{b10} developed a field delineation model focusing on France. However, work in \cite{b11} showed that this approach struggles to deliver accurate results in regions with smaller wheat fields, such as India. Recent publications such as \cite{b15} further discuss the importance of weakly supervised data in the crop monitoring approach.

Our wheat segmentation model proposed in~\cite{b1} is based on a transformer-based design that uses both time-series and spatial information. We relied on the Temporal Spatial Vision Transformer (TSViT), which was first trained on large datasets such as PASTIS \cite{b5} and then fine-tuned it using Parameter Efficient Fine-Tuning (PEFT) for our downstream task of winter wheat segmentation.

As input, the model takes multi-temporal Sentinel‑2 surface reflectance images covering the entire winter wheat growing season. Time-series data help the model learn how wheat fields change over time, while the spatial capabilities of the transformer let it recognize the shape and pattern of the fields. We trained the model on four years of Sentinel-2 data (2016–2019) and used 2020 as a holdout test set. 

\begin{figure}[ht]
\centering 
\includegraphics[width=0.4\textwidth,keepaspectratio]{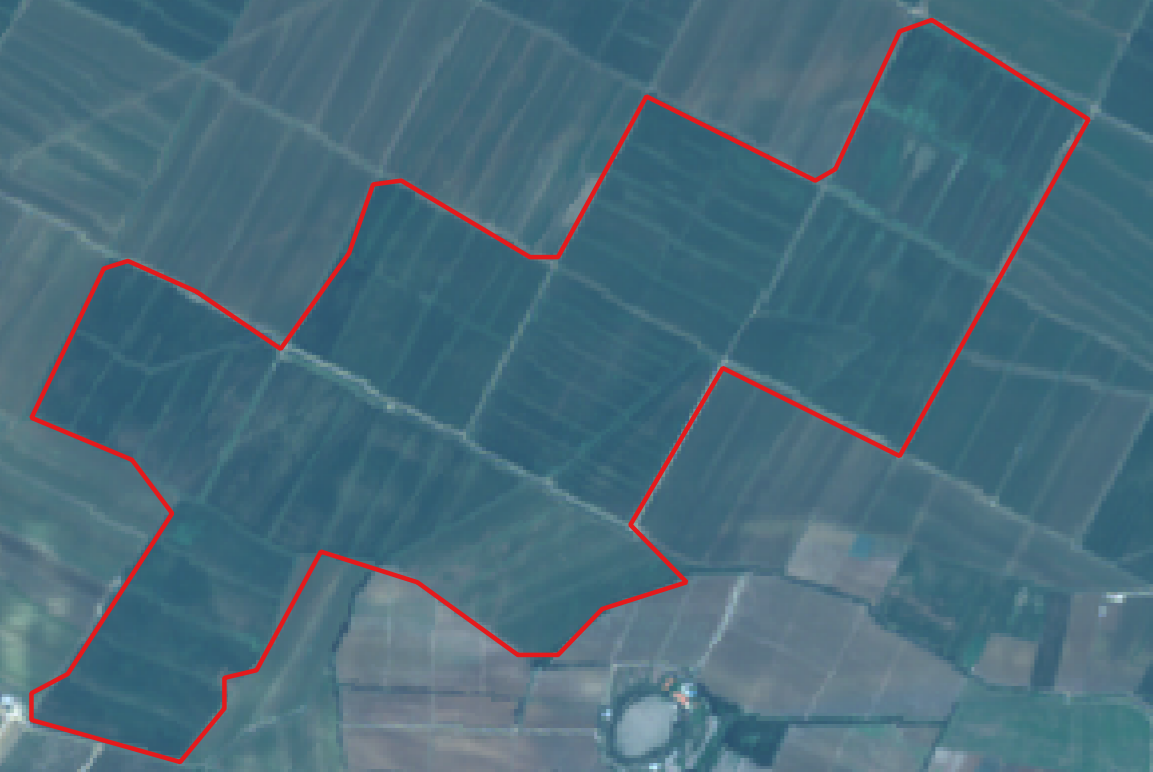}
\caption{Multiple wheat fields being incorrectly detected as one large field highlighted by the red polygon}
\label{largpoly}
\end{figure}

Although the wheat segmentation model proposed in~\cite{b1} performs well under controlled conditions, it encounters several obstacles and limitations, as discussed in the Introduction section earlier. 
The resulting masks lack the sharp contours needed to separate neighboring fields cleanly, an essential requirement for reliable instance delineation for operational agricultural monitoring. As shown in Figure \ref{largpoly}, a wheat-classified polygon (outlined in red) merges multiple wheat fields into a single polygon.

\begin{figure*}[ht]
    \centerline{\includegraphics[scale=0.4,keepaspectratio]{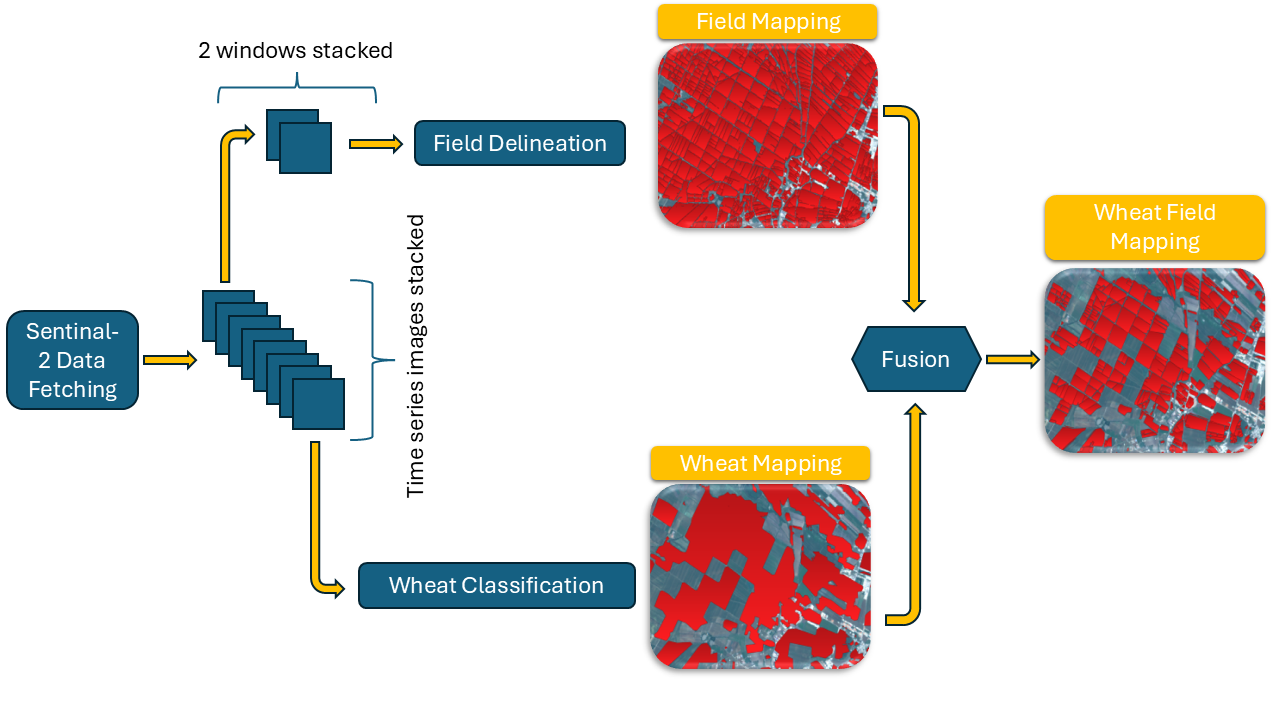}}
    \caption{ Overview of the proposed wheat field mapping pipeline. Sentinel-2 time-series images are processed through two parallel branches: (1) a \textbf{field delineation branch} for \textbf{parcel mapping} and (2) a \textbf{wheat classification} model for \textbf{wheat mapping}. The outputs from both branches are then fused to generate the final \textbf{wheat field mapping}.}
    \label{pipeline}
\end{figure*}

\section{Wheat Field Delineation}
To improve the spatial accuracy of our winter wheat segmentation, we integrated a dedicated field delineation model based on the open-source \texttt{ftw-baselines} framework~\cite{b2}. This model addresses limitations in spatial resolution and geometric inconsistencies by precisely extracting field boundaries. The model is trained on the Fields of The World (FTW) dataset, which contains 1.6 million parcel boundaries across 24 countries.

\subsection{Proposed Pipeline}
In Figure~\ref{pipeline}, we illustrate the complete proposed pipeline which is made up of the following steps:

\begin{itemize}
    \item \textbf{Sentinel-2 Data Fetcher:} Multi-temporal Sentinel-2 imagery is automatically fetched and preprocessed to serve as input for both the wheat classification and field delineation models. 
    
    \item \textbf{Wheat Classification:} A time-series stack of Sentinel-2 images covering the full growing season is fed into the wheat classification model, which outputs a mask identifying wheat-growing regions.
    
    \item \textbf{Field Delineation:} Two consecutive image windows are stacked and passed through the field delineation model, which produces masks representing individual agricultural field boundaries. This output is refined using our enhanced post-processing method.
    
    \item \textbf{Fusion Module:} The outputs of the field delineation and wheat classification models are combined by aligning the wheat predictions with the delineated field boundaries. Each detected crop field is labeled as wheat if more than 50\% of its area overlaps with wheat pixels. 
    
    \item \textbf{Noise Filtering and Regularization:} The resulting output is further cleaned by removing fields smaller than a threshold area. Field geometries are then simplified using the Ramer-Douglas-Peucker algorithm to produce smoother boundaries. The final product is a refined crop-specific map with accurate wheat segmentation constrained by real field shapes, making it suitable for large-scale agricultural monitoring.
\end{itemize}

\subsection{Enhanced Post-Processing}
Although the model in~\cite{b2} provides a solid foundation, it suffers from ambiguous boundaries and noise. Our post-processing techniques are designed to resolve these issues. 

Traditional argmax thresholding produces binary masks that often result in undersegmentation. To address this, we introduce \textbf{gradual thresholding}, a technique that sets pixel values below a defined threshold to zero while preserving the original scores of pixels above the threshold, creating a smoother and more informative segmentation mask. Formally, for a prediction score \(p\) and threshold \(T\), gradual thresholding is defined in Equation~\ref{eq:gradualthres}:

\begin{equation}
G(p; T) =
\begin{cases}
p, & p \ge T,\\
0, & p < T.
\end{cases}
\label{eq:gradualthres}
\end{equation}

\begin{figure*}[t]
  \centering
  \includegraphics[width=\textwidth]{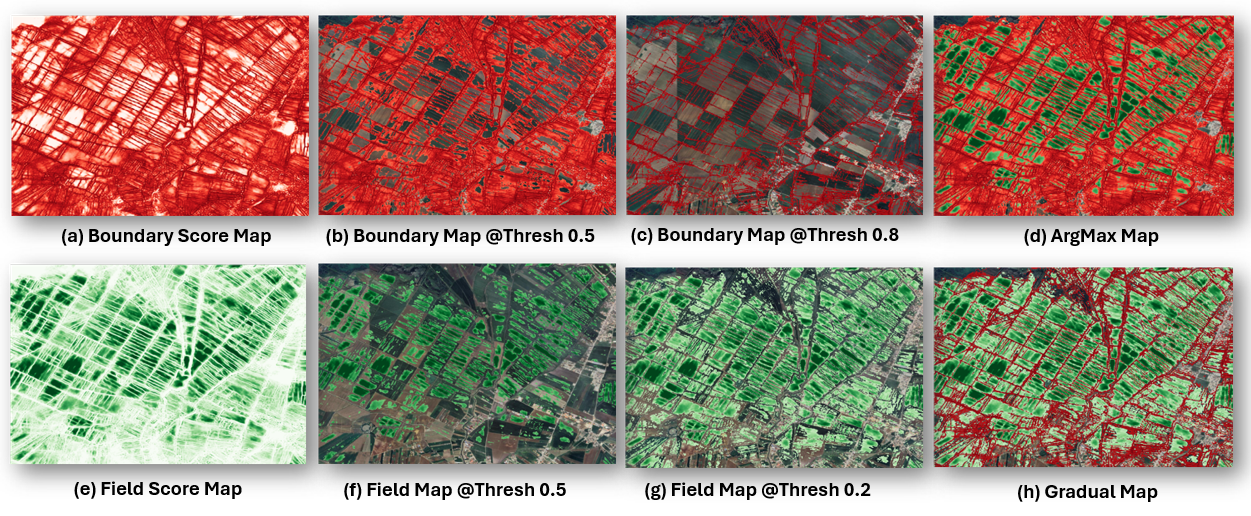}
  \caption{Comparison of boundary and field segmentation results under different thresholding strategies. Figures (a)–(c) show boundary score maps and their thresholded outputs at 0.5 and 0.8, respectively, while (d) illustrates the traditional argmax approach. Figures (e)–(g) present the field score map and its thresholded versions at 0.5 and 0.2, respectively and (h) demonstrates the final segmentation obtained using our gradual thresholding method.}
  \label{fig:all_maps}
\end{figure*}

We applied the following rules where these thresholds will be further discussed in Figure~\ref{fig:all_maps}:
\begin{itemize}
    \item \textbf{Strict Threshold for Boundary Mapping:} A threshold of \(T = 0.8\) is applied to boundary predictions to minimize uncertainty and accurately extract field edges.
    \item \textbf{Relaxed Threshold for Field Mapping:} A threshold of \(T = 0.2\) is used for field mapping to avoid under-segmentation and retain more complete field information.
\end{itemize}

After obtaining the gradual thresholded masks, we apply the following post-processing steps to refine the segmentation output:
\begin{itemize}
    \item \textbf{Creation of the Basins Mask:} We combine the gradual thresholded field mask with the inverted gradual thresholded boundary mask as shown in Equation~\ref{eq::basin}:

\begin{equation}
\textit{basins} = gradual\_fields \times (1.0 - gradual\_boundaries)
\label{eq::basin}
\end{equation}
  
    \item \textbf{Watershed Segmentation:} The basins mask, along with the gradual thresholded field mask, is fed into the watershed algorithm. The negative of the field probability scores is used to determine the basin depths, facilitating accurate instance segmentation.
    \item \textbf{Noise Filtering:} Additional filtering is applied to eliminate small spurious parcels, thereby reducing noise and improving overall segmentation quality.
\end{itemize}

\section{Performance Evaluation}
In Figure ~\ref{fig:all_maps}, We conducted several experiments to fine-tune the different threshold values used in the proposed pipeline.
\begin{itemize}
    \item \textbf{Boundary Confidence:} Figure~\ref{fig:all_maps} \textit{(a)} shows the boundaries confidence score map from the baseline field delineation model. 
    \begin{itemize}
        \item A threshold of 0.5 (Figure~\ref{fig:all_maps} \textit{(b)}) results in high uncertainty along the boundaries, leading to many fields being omitted.
        \item In contrast, thresholding at 0.8 (Figure~\ref{fig:all_maps} \textit{(c)}) retains most fields and produces better segmentation quality.
    \end{itemize}
    \item \textbf{Field Confidence:} Figure~\ref{fig:all_maps} \textit{(e)} displays the field confidence score map.
    \begin{itemize}
        \item When thresholded at 0.5 (Figure~\ref{fig:all_maps} \textit{(f)}), high uncertainty causes several fields to be removed.
        \item Thresholding at 0.2 (Figure~\ref{fig:all_maps} \textit{(g)}) preserves most fields and results in improved segmentation.
    \end{itemize}
    \item \textbf{ArgMax vs Gradual Thresholding:} Figure~\ref{fig:all_maps} \textit{(e)} illustrates the argmax approach, which led to undersegmented fields. In comparison, the gradual thresholding method (see Figure~\ref{fig:all_maps} \textit{(h)}) produces more accurate results, highlighting different fields.
\end{itemize}

In addition, we evaluated the performance of the proposed pipeline using in-house labeled ground truth data. Table \ref{tab:scores} summarizes the results for using Argmax and Gradual Thresholding. It is clear that Gradual Thresholding significantly outperforms Argmax in all metrics except precision. Although Argmax achieves slightly higher precision (0.9017 vs. 0.8379), this comes at the cost of a much lower recall (0.4615 vs. 0.7081), indicating that Argmax misses a large number of True Positive wheat pixels. Consequently, the F1-score for Argmax is substantially lower (0.6105) compared to Thresholding (0.7675), reflecting an imbalance between precision and recall. In terms of Intersection over Union (IoU) and Accuracy, Thresholding also demonstrates superior performance (0.6228 vs. 0.4394), suggesting more consistent agreement with the ground-truth delineations. These findings confirm that Thresholding provides a more balanced and reliable method for wheat field identification in our pipeline, making it a preferable choice. 

\begin{table}[ht] 
\centering 
\caption{Pipeline Performance Metrics for Argmax vs. Gradual Thresholding} 
\label{tab:scores} 
\begin{tabular}{|c|c|c|c|c|c|} 
    \hline \textbf{Method} & \textbf{IOU} & \textbf{Precision} & \textbf{Recall} & \textbf{F1-Score} & \textbf{Accuracy} \\
    \hline Argmax & 0.4394 & 0.9017 & 0.4615 & 0.6105 & 0.4394 \\
    \hline Thresholding & 0.6228 & 0.8379 & 0.7081 & 0.7675 & 0.6228 \\
    \hline 
\end{tabular} 
\end{table}

\section{Discussion and Future Work}
To understand the spatial and temporal dynamics of wheat cultivation over the past decade, we analyzed changes in wheat areas, focusing on expansion, abandonment, and land-use transitions.  Table \ref{tab:changeinarea} illustrates the instability of Lebanon's economic and social landscape. Wheat cultivation in Lebanon experienced considerable fluctuations between 2016 and 2024, with peaks in 2019 and 2023 (same for number of fields), and significant drops in 2020 and 2024. The sharp decline in wheat cultivation from 2019 to 2020 reflects the impact of global events, such as the COVID-19 pandemic. More specifically, this substantial decrease between 2019 and 2020 strongly suggests the impact of the severe financial crisis and economic meltdown that hit the country at that time. Again in 2024, political turmoil and armed conflicts left a major impact on the wheat area. The resulting wheat maps can be publicly accessed at https://geogroup.ai/catalogue/\#wheat

\begin{table}[ht]
    \centering
    \caption{Wheat Field Area (km²), Fields Count, and Vast Wheat Area (km²) in Lebanon, 2016--2024}
    \label{tab:changeinarea}
    \begin{tabular}{|c|c|c|c|}
        \hline
        \textbf{Year} & \textbf{Wheat Fields Surface} & \textbf{Number of Fields} & \textbf{Vast Wheat Areas} \\
        \hline 
        2016	& 247.6 & 15,691 &	75.6   \\
        2017	& 252.4 & 12,327 &	136.8  \\
        2018	& 241.9 & 15,635 &	157.4  \\
        2019	& 365.6 & 20,914 &	262.1  \\
        2020	& 203.0 & 12,113 &	50.9   \\
        2021	& 283.8 & 16,805 &	116.4  \\
        2022	& 263.1 & 15,049 &	78.9   \\
        2023	& 315.2 & 17,867 &	121.5  \\
        2024	& 199.1 & 11,638 &	63.7   \\
        \hline
    \end{tabular}
\end{table}
Conceptually, \emph{Vast Wheat Areas} quantify wheat outside parcel-like fields (e.g., diffuse plantings) thus beyond FTW's ``field'' ontology; elevated values mostly signal omission.


\begin{figure*}[t]
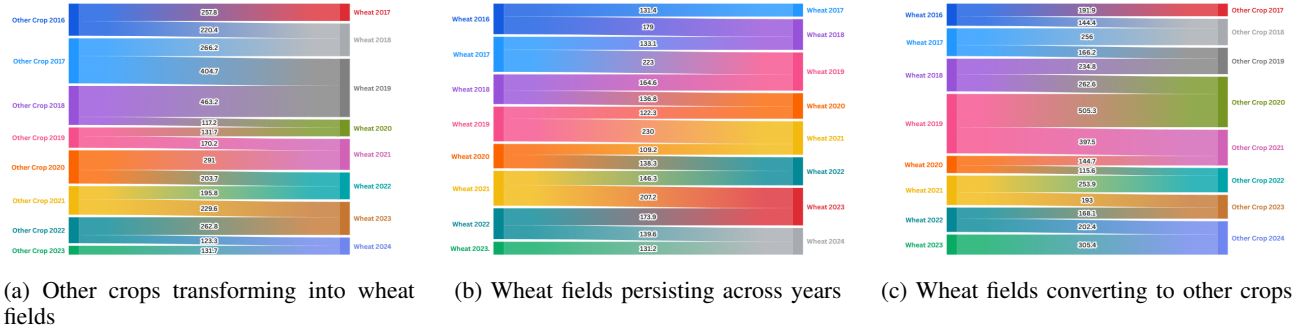

    \centering
    \subcaptionbox{Other crops transforming into wheat fields\label{fig:sub1}}[0.3\textwidth]{%
        \includegraphics[width=\linewidth]{Figures/new-wheat-flow-2.png}%
    }%
    \hspace{0.02\textwidth}%
    \subcaptionbox{Wheat fields persisting across years\label{fig:sub2}}[0.3\textwidth]{%
        \includegraphics[width=\linewidth]{Figures/wheat-rotation-2.png}%
    }%
    \hspace{0.02\textwidth}%
    \subcaptionbox{Wheat fields converting to other crops\label{fig:sub3}}[0.3\textwidth]{%
        \includegraphics[width=\linewidth]{Figures/abandoned-wheat-flow-2.png}%
    }%
    \caption{Sankey diagrams illustrating the transitions in crop field areas between annual and bi-annual years. (a) Transformation of other crop areas into wheat fields; (b) Persistence of wheat areas across years; (c) Conversion of wheat fields into other crops.}
    \label{fig:multi-sankey}
\end{figure*}

In addition to the numerical analysis presented in Table \ref{tab:changeinarea}, we use a Sankey diagram, a visual method to represent flows between different states to understand crop rotation practices. Figure \ref{fig:multi-sankey}-(a) shows a significant peak in 2019 (indicated by the thickest flow), suggesting strong incentives or favorable conditions for new wheat areas. The lowest annual new wheat area occurred in 2024, which corresponds to armed conflict and labor availability.

Figure \ref{fig:multi-sankey}-(b) shows that the flow of intersected areas between biannual years is greater than between annual years, suggesting a crop rotation scheme of Lebanese farmers over a two-year gap rather than one. 

Figure \ref{fig:multi-sankey}-(c) shows that the highest abandonment rate was in 2019-2020 ($505.3 km^2$), closely correlated with the significant decline in the total wheat area in 2020, possibly due to economic distress. The lowest abandonment was $144.7 km^2$ (2020-2021), suggesting a partial recovery or stabilization after significant disruptions.

According to USDA \cite{b3}, the wheat yield in Lebanon for 2023 was 3.5 tons per hectare. In contrast, another study \cite{b4} reports a yield of 2.5 tons per hectare for the same year. 
Our model currently maps wheat fields without differentiating between different breeds and irrigation methods. In Lebanon, there are two main breeds of wheat: durum and soft wheat, each with its own yield characteristics.
In addition, currently we do not distinguish between rainfed fields and irrigated lands. Consequently, we cannot accurately determine national wheat production, as each field yield varies depending on its type and irrigation practices. This highlights a new research challenge: classify rainfed and irrigated fields and classify durum and soft wheat parcels. Incorporating these labels would enable us to report precise yield estimations. 


\section*{Acknowledgments}
This work was partially funded through ESA NoR 4228A5.


\bibliographystyle{unsrt} 
\bibliography{refs} 

\begin{thebibliography}{10}

\bibitem{b1}
Mohamad~Hasan Zahweh, Hasan Nasrallah, Mustafa Shukor, Ghaleb Faour, and Ali~J Ghandour.
\newblock Empirical study of peft techniques for winter-wheat segmentation.
\newblock {\em Environmental Sciences Proceedings}, 29(1):50, 2023.

\bibitem{b2}
Hannah Kerner, Snehal Chaudhari, Aninda Ghosh, Caleb Robinson, Adeel Ahmad, Eddie Choi, Nathan Jacobs, Chris Holmes, Matthias Mohr, Rahul Dodhia, et~al.
\newblock Fields of the world: A machine learning benchmark dataset for global agricultural field boundary segmentation.
\newblock {\em arXiv preprint arXiv:2409.16252}, 2024.

\bibitem{b9}
Rose M~Rustowicz, Robin Cheong, Lijing Wang, Stefano Ermon, Marshall Burke, and David Lobell.
\newblock Semantic segmentation of crop type in africa: A novel dataset and analysis of deep learning methods.
\newblock In {\em Proceedings of the IEEE/cvf conference on computer vision and pattern recognition workshops}, pages 75--82, 2019.

\bibitem{b6}
Vivien Sainte~Fare Garnot, Loic Landrieu, and Nesrine Chehata.
\newblock Multi-modal temporal attention models for crop mapping from satellite time series.
\newblock {\em ISPRS Journal of Photogrammetry and Remote Sensing}, 187:294--305, 2022.

\bibitem{b7}
Michail Tarasiou, Erik Chavez, and Stefanos Zafeiriou.
\newblock Vits for sits: Vision transformers for satellite image time series.
\newblock In {\em Proceedings of the IEEE/CVF Conference on Computer Vision and Pattern Recognition}, pages 10418--10428, 2023.

\bibitem{b12}
Gabriel Tseng, Ivan Zvonkov, Catherine~Lilian Nakalembe, and Hannah Kerner.
\newblock Cropharvest: A global dataset for crop-type classification.
\newblock In {\em Thirty-fifth Conference on Neural Information Processing Systems Datasets and Benchmarks Track (Round 2)}, 2021.

\bibitem{b13}
United States~Department of~Agriculture (USDA) Foreign Agricultural~Service.
\newblock Automated delineation of agricultural field boundaries from sentinel-2 images using recurrent residual u-net.
\newblock {\em International Journal of Applied Earth Observation and Geoinformation}, 2021.

\bibitem{b14}
Dujuan Zhang, Yaozhong Pan, Jinshui Zhang, Tangao Hu, Jianhua Zhao, Nan Li, and Qiong Chen.
\newblock A generalized approach based on convolutional neural networks for large area cropland mapping at very high resolution.
\newblock {\em Remote Sensing of Environment}, 247:111912, 2020.

\bibitem{b10}
Fran{\c{c}}ois Waldner, Foivos~I Diakogiannis, Kathryn Batchelor, Michael Ciccotosto-Camp, Elizabeth Cooper-Williams, Chris Herrmann, Gonzalo Mata, and Andrew Toovey.
\newblock Detect, consolidate, delineate: Scalable mapping of field boundaries using satellite images.
\newblock {\em Remote sensing}, 13(11):2197, 2021.

\bibitem{b11}
Sherrie Wang, Fran{\c{c}}ois Waldner, and David~B Lobell.
\newblock Unlocking large-scale crop field delineation in smallholder farming systems with transfer learning and weak supervision.
\newblock {\em Remote Sensing}, 14(22):5738, 2022.

\bibitem{b15}
KC~Sumesh, Jagannath Aryal, and Dongryeol Ryu.
\newblock A novel architecture for automated delineation of the agricultural fields using partial training data in remote sensing images.
\newblock {\em Computers and Electronics in Agriculture}, 234:110265, 2025.

\bibitem{b5}
Vivien Sainte~Fare Garnot and Loic Landrieu.
\newblock Panoptic segmentation of satellite image time series with convolutional temporal attention networks.
\newblock In {\em Proceedings of the IEEE/CVF International Conference on Computer Vision}, pages 4872--4881, 2021.

\bibitem{b3}
{L}ebanon {W}heat {A}rea, {Y}ield and {P}roduction --- ipad.fas.usda.gov.
\newblock \url{https://ipad.fas.usda.gov/countrysummary/default.aspx?id=LE&crop=Wheat}.
\newblock [Accessed 05-04-2025].

\bibitem{b4}
{W}heat yields --- ourworldindata.org.
\newblock \url{https://ourworldindata.org/grapher/wheat-yields?tab=chart&country=~LBN}.
\newblock [Accessed 05-04-2025].

\end{thebibliography}

\vspace{12pt}

\end{document}